\documentclass[11pt]{article}

\usepackage[final]{acl}

\usepackage{times}
\usepackage{latexsym}

\usepackage[T1]{fontenc}

\usepackage[utf8]{inputenc}

\usepackage{microtype}

\usepackage{inconsolata}

\usepackage{graphicx}
\usepackage{amssymb}
\usepackage{pifont}
\usepackage{booktabs}
\usepackage{multirow}
\usepackage{makecell}
\usepackage{xcolor}
\usepackage{subfig}
\usepackage[most]{tcolorbox}
\usepackage[ruled,vlined]{algorithm2e}
\newcommand{\methodname}{MaskTab}
\newcommand{\method}{\mbox{\methodname}\xspace}

\title{MaskTab: Scalable Masked Tabular Pretraining with Scaling Laws and Distillation for Industrial Classification}

\author{
  Bo Zheng\textsuperscript{1,2}\thanks{These authors contributed equally to this work.},
  Yudong Chen\textsuperscript{2}\footnotemark[1],
  Zihua Xiong\textsuperscript{2},
  Shuai Fang\textsuperscript{2},\\
  \textbf{Peidong He}\textsuperscript{2},
  \textbf{Yang Yang}\textsuperscript{1},
  \textbf{Sheng Guo}\textsuperscript{2}\thanks{Corresponding author.} \\[0.5em]
  \textsuperscript{1}Zhejiang University \quad
  \textsuperscript{2}MyBank, Ant Group \\[0.3em]
  \texttt{guosheng1001@gmail.com} \\
}

\begin{document}
\maketitle

\begin{abstract}
Tabular data forms the backbone of high-stakes decision systems in finance, healthcare, and beyond. Yet industrial tabular datasets are inherently difficult: high-dimensional, riddled with missing entries, and rarely labeled at scale. While foundation models have revolutionized vision and language, tabular learning still leans on handcrafted features and lacks a general self-supervised framework.
We present MaskTab, a unified pre-training framework designed specifically for industrial-scale tabular data. MaskTab encodes missing values via dedicated learnable tokens, enabling the model to distinguish structural absence from random dropout. It jointly optimizes a hybrid supervised pre-training scheme—utilizing a twin-path architecture to reconcile masked reconstruction with task-specific supervision—and an MoE-augmented loss that adaptively routes features through specialized subnetworks.
On industrial-scale benchmarks, it achieves +5.04\% AUC and +8.28\% KS over prior art under rigorous scaling.
Moreover, its representations distill effectively into lightweight models, yielding +2.55\% AUC and +4.85\% KS under strict latency and interpretability constraints, while improving robustness to distribution shifts. Our work demonstrates that tabular data admits a foundation-model treatment—when its structural idiosyncrasies are respected.

\end{abstract}
\section{Introduction}\label{sec:intro}

\begin{figure}[t]
\begin{center}
\includegraphics[width=1.0\linewidth]{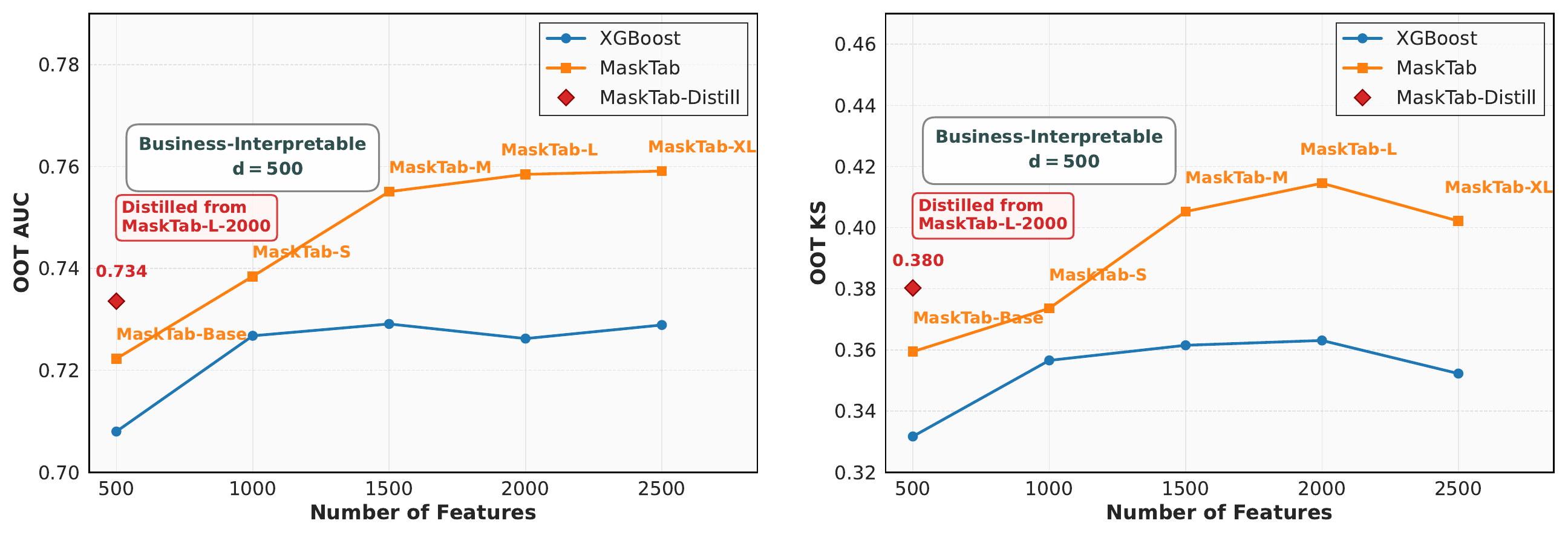}
\end{center}
\caption{Overall Model Performance. Average monthly AUC (left) and KS (right) on OOT data for \method and XGBoost~\citep{xgboost}, as the number of features increases from 500 to 2500. \method consistently outperforms XGBoost, with larger gains at higher feature dimensions. \texttt{MaskTab-Distill} further transfers the benefits of \texttt{MaskTab-L} to a 500-feature, interpretable setting, outperforming both \texttt{MaskTab-Base} and XGBoost.}
\label{fig:ks_comp}
\end{figure}

Tabular data lies at the core of high-stakes automated decision systems---credit scoring, risk prediction, and fraud detection---where model reliability directly impacts financial outcomes. Yet despite its operational centrality, tabular representation learning has seen little of the paradigm shift brought by self-supervised pretraining in other domains. Real-world industrial tables are large-scale, sparse, and only weakly labeled; more critically, missing values are often informative rather than ignorable, reflecting  underlying data-generating processes. Standard pipelines, however, treat such structure as noise: features with high missing rates are discarded, continuous variables are naively imputed, and categorical fields are reduced to static encodings. The result is a loss of signal before learning even begins.


Current practice reflects this gap. Gradient-boosted Decision trees (GBDTs) like XGBoost \citep{xgboost} dominate production pipelines not because they are optimal, but because they tolerate raw inputs with minimal preprocessing. Deep alternatives such as FT-Transformer \citep{gorishniy2021revisiting} improve expressivity but still treat tabular data as a passive collection of columns, ignoring the semantics of absence and scaling poorly beyond a few hundred features. Crucially, none offer a self-supervised pretraining regime that leverages unlabeled data at scale—a missed opportunity given the abundance of unlabeled records in enterprise settings.

We contend that tabular data can support foundation models, provided the architecture respects its statistical idiosyncrasies. To this end, we introduce \method, a pretraining framework built on three design principles: (1)\textit{Missingness as signal:} Instead of imputing or dropping incomplete entries, MaskTab represents missing values with dedicated learnable tokens, allowing the model to distinguish informative absence from noise. (2) \textit{Unified pretraining:} A Siamese twin-path architecture jointly optimizes masked reconstruction and supervised objectives, enabling seamless adaptation from pretraining to downstream tasks—even with scarce labels. (3) \textit{Scalable interaction modeling:} A Mixture-of-Experts (MoE) reconstruction head adaptively routes high-dimensional features through specialized subnetworks, capturing complex interactions without combinatorial explosion.

As shown in Figure~\ref{fig:ks_comp}, \method consistently outperforms XGBoost across feature scales (500–2500), with gains widening as dimensionality grows. Notably, a distilled variant—constrained to 500 interpretable features—surpasses both XGBoost and our base model, demonstrating that rich representations can be compressed without sacrificing performance or robustness to temporal shifts.

Our results establish that tabular data admits a foundation-model treatment when representation learning is grounded in its operational reality. 
\method sets a new state-of-the-art benchmark on industrial datasets, achieving a relative improvement of 5.04\% in AUC\footnote{Area Under the ROC Curve, measures classification performance.} 
 and 8.28\% in KS\footnote{Kolmogorov--Smirnov statistic, measures class separation in risk modeling.} over existing baselines. 
Furthermore, to address the stringent latency and regulatory interpretability requirements of production environments, we develop a knowledge distillation pipeline that transfers knowledge to a business-audited feature subset.
The resulting lightweight student models retain the powerful semantics of the teacher, yielding a +2.55\% AUC and +4.85\% KS gain over traditional models while exhibiting superior robustness to temporal distribution shifts. More broadly, our results suggest that tabular data should no longer be regarded as a “solved” engineering problem, but as an open frontier for representation learning.

\section{Related Work}\label{sec:related_work}

\noindent\textbf{Tabular Models.}
In tabular data modeling, tree-based ensembles such as XGBoost \citep{xgboost}, LightGBM \citep{lightgbm}, and CatBoost \citep{catboost} have long been established as strong baselines due to their robustness, interpretability, and performance on structured data. With advancements in deep learning, research has shifted towards deep tabular models like ResNet \citep{gorishniy2021revisiting}, SNN \citep{SNN}, and DCNv2 \citep{dcnv2}. More recently, Transformer-based approaches have gained prominence by treating individual features as tokens and embedding feature values into high-dimensional spaces. Notable methods in this line include TransTab \citep{wang2022transtab}, FT-Transformer \citep{gorishniy2021revisiting}, which leverage self-attention to capture complex feature interactions. Going beyond conventional task-specific training, TabPFN \citep{grinsztajn2026tabpfn25advancingstateart} introduces a tabular foundation model pretrained on millions of synthetic datasets, enabling in-context learning and approximate Bayesian inference for small-scale classification and regression tasks without additional training. Further studies have proposed improved embedding techniques for different data types, such as enhanced representations for numerical features \citep{mlp-plr} and pre-trained BERT embeddings for categorical/textual attributes \citep{cm2}, respectively. Another line of work adopts neighborhood-based retrieval, where test samples are compared against the training set to identify nearest neighbors for prediction, as seen in TabR \citep{tabr}.

\noindent\textbf{Scaling Law.} Scaling laws characterize the predictable relationship between model performance and key scaling variables such as model size, dataset size, and computational budget. Early work by \citep{kaplan2020scaling} established power-law scaling in language models, later refined by DeepMind's Chinchilla laws \citep{hoffmann2022training} and Gopher study \citep{rae2021scaling}. Similar trends have been observed in vision \citep{zhai2022scaling} and domain-specific settings such as recommender systems \citep{lai2025exploring} and time series modeling \citep{yao2024towards}. For tabular data, \citet{ma2024tabdpt} recently initiated scaling studies, yet most efforts focus on model and data scaling, overlooking feature dimension scaling and unlabeled data utilization—especially under realistic conditions with heterogeneous types and systematic missingness.

\noindent\textbf{Knowledge Distillation.}
Knowledge distillation (KD) is a widely used technique for compressing large models into smaller, efficient variants while preserving performance. \citep{hinton2015distilling} established the paradigm of using softened teacher logits ("soft targets") as training labels for the student, capturing dark knowledge via KL divergence loss. In tabular data, Knowledge Distillation has been used to transfer expertise from dominant tree ensembles like GBDTs into neural networks, as seen in DeepGBM \citep{ke2019deepgbm}, which leverages tree outputs to guide the student model.
More recently, \citep{wang2022transtab} applied distillation in their TransTab framework, where a large transformer-based teacher model pre-trained on diverse tabular datasets transfers knowledge to a smaller student model for improved generalization on downstream tasks. However, distilling tabular models—especially those handling missing values and heterogeneous groups—under strict latency and memory constraints remains under-explored. 
\section{MaskTab}\label{sec:method}

\subsection{Task Definition}
We study tabular classification with a labeled set and a much larger unlabeled set:

\begin{equation}
\begin{split}
\mathcal{D}&=\mathcal{D}_{\text{sup}}\cup\mathcal{D}_{\text{unsup}},\quad
\mathcal{D}_{\text{sup}}=\{(\mathbf{x}_i,y_i)\}_{i=1}^N,\\
\mathcal{D}_{\text{unsup}}&=\{\mathbf{x}_j\}_{j=1}^M,\quad M\gg N.
\end{split}
\end{equation}

Each instance $\mathbf{x}=(v_1,\ldots,v_d)$ contains heterogeneous features (numeric, categorical, text, or missing). We use binary classification as a running example, with labels \(y\in\{0,1\}\), however, \method naturally extends to \(n\)-way classification and regression.

\subsection{Baseline: Transformer Encoder for Heterogeneous Tables}
We adopt an encoder-only Transformer as the backbone for modeling feature interactions in heterogeneous tabular data. For each feature $k$, we construct a token embedding by combining a feature-name embedding and a feature-value embedding:
\begin{equation}
\mathbf{h}_k=\text{LayerNorm}\big(\mathbf{e}^{\text{name}}_k+\phi(v_k)\big)\in\mathbb{R}^h.
\end{equation}
Feature names are encoded once using a frozen pre-trained language encoder (BERT \citep{devlin2018bert}) with pooling to obtain $\mathbf{e}^{\text{name}}_k$. The value encoder $\phi(\cdot)$ is type-specific: a linear projection for numerical values, an embedding lookup for categorical values, a frozen text encoder for textual values, and a shared learnable embedding for missing values.

\begin{figure*}[!htb]
\begin{center}
\includegraphics[width=0.9\linewidth]{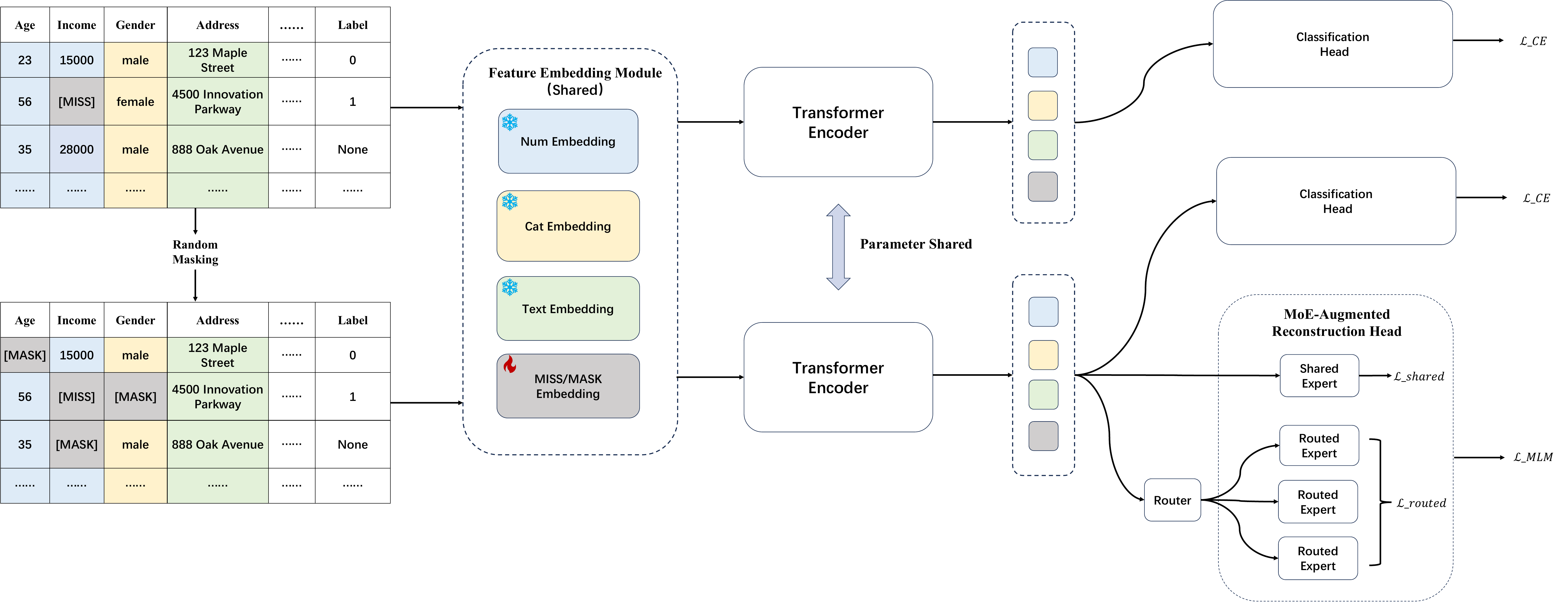}
\end{center}
\caption{MaskTab encodes high-dimensional tabular data with learnable tokens, then hybrid pre-trains a tabular encoder with a twin-path objective: MoE-based masked reconstruction on unlabeled data and supervised learning on labeled data.}
\label{fig:main}
\end{figure*}

Stacking all feature tokens yields $\mathbf{H}\in\mathbb{R}^{d\times h}$, which is fed into a Transformer to capture cross-feature dependencies:
\begin{equation}
\mathbf{Z}=\text{Transformer}(\mathbf{H}).
\end{equation}
We obtain a row-level representation by mean pooling and apply a linear classification head to predict $\hat{y}$. The model is trained on $\mathcal{D}_{\text{sup}}$ with binary cross-entropy loss. Building on this, we introduce \method, as shown in Figure~\ref{fig:main}.


\subsection{Learnable Tokens for Masked and Real Missing Values}
\label{sec:dual-missing}

We propose a self-supervised objective that treats missingness as a learnable signal and unifies \emph{synthetic masking} and \emph{natural missing values} with dedicated tokens.
Concretely, we use two special symbols: \texttt{[MASK]} for randomly masked observed entries ($\mathcal{M}$) during pre-training and \texttt{[MISS]} for naturally missing entries ($v_k=\bot$).
Their embeddings are initialized identically with a shared learnable vector $\mathbf{m}\in\mathbb{R}^h$, referred to as the \emph{mask embedding}, which reduces the mismatch between pre-training and real missingness:
\begin{align}
\phi(v_k)
&=
\begin{cases}
\mathbf{e}_{\texttt{[MASK]}}, & k\in\mathcal{M},\\
\mathbf{e}_{\texttt{[MISS]}}, & v_k=\bot,\\
\text{standard encoding}, & \text{otherwise},
\end{cases}
\notag\\
\mathbf{e}_{\texttt{[MASK]}}
&=
\mathbf{e}_{\texttt{[MISS]}}
=
\mathbf{m}
\quad \text{(at init)} .
\end{align}

\paragraph{Masked value reconstruction.}
Given an unlabeled instance $\mathbf{x}=(v_1,\ldots,v_d)$, we sample a subset of observed features $\mathcal{M}$ and replace them with \texttt{[MASK]} to obtain $\widetilde{\mathbf{x}}$.
After encoding, $\mathbf{Z}=\text{Transformer}_\Theta(\phi(\widetilde{\mathbf{x}}))$, we predict each masked value from its contextual representation:
\begin{equation}
\mathcal{L}_{\text{MLM}}=\sum_{k\in\mathcal{M}} \ell\!\left(g(\mathbf{z}_k),\, v_k\right),
\end{equation}
where $g(\cdot)$ is a lightweight prediction head shared across features and $\ell$ is type-aware (e.g., MSE for numerical features and cross-entropy/contrastive loss for categorical or textual features).
This objective encourages the encoder to exploit both feature co-occurrence patterns and missingness structure.


\subsection{Hybrid Supervised Pre-Training with Twin Networks}
\label{sec:mixed-supervision}

Conventional two-stage pre-training (self-supervised pre-train \(\rightarrow\) supervised fine-tune) is often suboptimal for industrial tabular learning. It (i) requires separate hyperparameter schedules across stages, and (ii) suffers from \emph{objective misalignment}: the checkpoint minimizing the reconstruction loss does not necessarily optimize downstream classification, even when substantial labeled data are available.

We therefore adopt a \emph{hybrid supervised} objective that trains on labeled and unlabeled data in a single, unified procedure. For labeled samples \((\mathbf{x}, y)\), we jointly optimize masked reconstruction and classification:
\begin{equation}
\mathcal{L}_{\text{hybrid}}^{(\text{sup})}
= \lambda \, \mathcal{L}_{\text{MLM}} + (1-\lambda)\, \mathcal{L}_{\text{CE}}, \quad \lambda \in [0,1],
\end{equation}
where \(\mathcal{L}_{\text{MLM}}\) is the reconstruction loss in Sec.~\ref{sec:dual-missing} and \(\mathcal{L}_{\text{CE}}\) is the binary cross-entropy loss. For unlabeled samples \(\mathbf{x}\), we apply only the self-supervised objective:
\begin{equation}
\mathcal{L}_{\text{hybrid}}^{(\text{self-sup})}=\mathcal{L}_{\text{MLM}}.
\end{equation}
The overall objective averages over labeled and unlabeled sets:
\begin{equation}
\begin{aligned}
\mathcal{L}_{\text{hybrid}}
&=\frac{1}{|\mathcal{D}_{\text{sup}}|}
\sum_{(\mathbf{x},y)\in \mathcal{D}_{\text{sup}}}
\mathcal{L}_{\text{hybrid}}^{(\text{sup})} \\
&\quad+\frac{1}{|\mathcal{D}_{\text{unsup}}|}
\sum_{\mathbf{x}\in \mathcal{D}_{\text{unsup}}}
\mathcal{L}_{\text{hybrid}}^{(\text{self-sup})}.
\end{aligned}
\end{equation}
We use the same formulation throughout training, avoiding an explicit pre-train/fine-tune boundary.

\paragraph{Twin-path training to avoid masking-induced shift.}
Directly feeding masked inputs to the classifier trains \(P(y\mid \widetilde{\mathbf{x}})\), introducing a train--test mismatch because inference uses the natural missing pattern in \(\mathbf{x}\).
To eliminate this shift, for each labeled instance we employ two parallel paths with shared Transformer parameters \(\Theta\) (Fig.~\ref{fig:main}):

\noindent\textbf{Reconstruction path.} We apply masking to obtain \(\widetilde{\mathbf{x}}\) and compute
\begin{equation}
\begin{aligned}
&\mathbf{Z}^{(\text{MLM})} = \text{Transformer}_{\Theta}(\phi(\widetilde{\mathbf{x}})), \\
& \qquad \mathcal{L}_{\text{CE}}\!\left(\mathbf{z}^{(\text{MLM})}\right), \qquad \mathcal{L}_{\text{MLM}}(\mathbf{Z}^{(\text{MLM})}).
\end{aligned}
\end{equation}

\noindent\textbf{Classification path.} We keep the original input \(\mathbf{x}\) and compute
\begin{equation}
\begin{aligned}
&\mathbf{z}^{(\text{CLS})}
= \text{Pooling}\!\left(
\text{Transformer}_{\Theta}\big(\phi(\mathbf{x})\big)
\right), \\
&\qquad \mathcal{L}_{\text{CE}}\!\left(\mathbf{z}^{(\text{CLS})}\right).
\end{aligned}
\end{equation}

Sharing \(\Theta\) transfers dependency/missingness knowledge learned by reconstruction to the classifier, while ensuring the classifier is trained on the true input distribution \(P(y\mid \mathbf{x})\).

\paragraph{Adaptive masking.}
To stabilize training for inherently sparse instances, we reduce corruption as the original missing ratio increases. Let
\(\eta(\mathbf{x})=\frac{1}{d}\sum_{k=1}^d \mathbb{I}(v_k=\bot)\) be the missing ratio and \(\eta_{\max}\) the maximum over the dataset. We set the instance-wise masking rate as
\begin{equation}
r_{\text{mask}}(\mathbf{x})
= r_{\max}\left(1-\frac{\eta(\mathbf{x})}{\eta_{\max}}\right)^{\alpha},
\end{equation}
where \(r_{\max}\) is the maximal masking ratio and \(\alpha>0\) controls sensitivity.
This preserves information for highly-missing samples while still enforcing strong corruption for complete ones.

\subsection{Scalable Mixture-of-Experts Feature Grouping}
\label{sec:MoE}

A single linear projection for masked feature reconstruction becomes a bottleneck as the number of tabular features grows. We therefore replace the monolithic MLM head with a lightweight Mixture-of-Experts (MoE) projection layer, inspired by recent MoE LLMs (e.g., DeepSeekMoE; \citealp{dai2024deepseekmoe}). The key idea is to \emph{route features to specialized experts}: each feature token selects a small subset of experts for reconstruction, increasing capacity while keeping per-token computation bounded. Concretely, we combine (i) one \emph{shared} expert, applied to all masked features, and (ii) \(K_r\) \emph{routed} experts with top-\(K_a\) gating. The MLM objective becomes
\begin{equation}
\mathcal{L}_{\text{MLM}}=\alpha\,\mathcal{L}_{\text{Shared}}+\beta\,\mathcal{L}_{\text{Routed}},
\end{equation}
where \(\alpha,\beta\) weight the shared and routed terms, and \(\mathcal{M}\) denotes masked positions. The shared expert loss is
\begin{equation}
\mathcal{L}_{\text{Shared}}
=\sum_{k\in\mathcal{M}}
\ell\!\left(\mathbf{w}_{0}\mathbf{z}_k+\mathbf{b}_{0},\, v_k\right),
\end{equation}
and the routed expert loss is
\begin{equation}
\mathcal{L}_{\text{Routed}}
=\sum_{k\in\mathcal{M}}
\ell\!\left(\sum_{i=1}^{K_r} g_{i,k}\big(\mathbf{w}_{i}\mathbf{z}_k+\mathbf{b}_{i}\big),\, v_k\right).
\end{equation}
For feature \(k\), the gate \(g_{i,k}\) keeps only the top-\(K_a\) routed experts:
\begin{equation}
g_{i,k}=
\begin{cases}
s_{i,k}, & s_{i,k}\in \text{TopK}(\{s_{j,k}\}_{j=1}^{K_r},\,K_a),\\
0, & \text{otherwise},
\end{cases}
\end{equation}
with routing scores computed by centroid matching
\begin{equation}
s_{i,k}=\text{Softmax}_{i}\!\left(\mathbf{z}_k^{\top}\mathbf{e}_i\right),
\end{equation}
where \(\mathbf{z}_k\) is the token representation, \(\{\mathbf{e}_i\}_{i=1}^{K_r}\) are learnable expert centroids, and \((\mathbf{w}_i,\mathbf{b}_i)\) are expert-specific output projections. This design encourages adaptive feature grouping and expert specialization, substantially improving reconstruction capacity without introducing heavy expert MLPs.


\subsection{Task-Specialized Distillation with Interpretable Features}
\label{sec:distillation}

Despite their formidable capabilities, pre-trained models are often prohibitively large and utilize an extensive set of features, rendering them unsuitable for scenarios like credit scoring that demand low latency and high interpretability. To derive a practical model, we distill its knowledge into a smaller version of \method that uses only a predefined, interpretable feature set.

The student model is trained by employing a embedding alignment loss to match the teacher's output distribution, combined with a hybrid-supervision loss (detailed in Section \ref{sec:mixed-supervision}). Specifically, to align the intermediate representations, the student's final hidden state $\mathbf{z}^{\text{(CLS)}}$ is first projected into the teacher's embedding space using a linear layer that increases its dimensionality to match that of the teacher's representation
 using a linear layer that increases its dimensionality from $d_{s}$ to $d_{t}$:

\begin{equation}
\mathbf{z}_{a} = 
\mathbf{w}_{a}\mathbf{z}^{\text{(CLS)}} + \mathbf{b}_{a},
\end{equation}

where $\mathbf{w}_{a} \in \mathbb{R}^{d_{t} \times d_{s}}$ and $\mathbf{b}_{a} \in \mathbb{R}^{d_{t}}$ are learnable parameters. We then minimize a representation loss $\mathcal{L}_{\text{align}}$ defined as the negative cosine similarity between the projected student representation $\mathbf{z}_{a}$ and the embedding $\mathbf{e}_{t} \in \mathbb{R}^{d_{t}}$ generated by the \texttt{MaskTab-L} teacher: 
\begin{equation}
\mathcal{L}_{\text{align}} = 1 - \cos(\mathbf{z}_{a}, \mathbf{e}_t) = 1 - \frac{\mathbf{z}_{a} \cdot \mathbf{e}_t}{|\mathbf{z}_{a}| |\mathbf{e}_t|}.
\end{equation}

Therefore, the overall training objective for the student model is a weighted combination of multiple losses: 

\begin{equation}
\begin{aligned}
\mathcal{L}_{\text{hybrid}}^{(\text{sup})} =
& \lambda_1 \cdot \mathcal{L}_{\text{MLM}} + \lambda_2 \cdot \mathcal{L}_{\text{CE}} + \lambda_3 \cdot \mathcal{L}_{\text{align}}, \\
& \quad \lambda_1 + \lambda_2 + \lambda_3 = 1.
\end{aligned}
\end{equation}
\section{Experiments}\label{sec:exp}

\begin{table*}[t]
\centering
\caption{
Comparing \method with current Gradient Boosted Decision Tree (GBDT) methods and deep tabular models using the TabReD benchmark. The entries highlighted in bold indicate the best-performing models.
}
\label{tab:public_dataset_performance}
\resizebox{\linewidth}{!}{
\begin{tabular}{@{}lccccccccc@{}}

\toprule
\multicolumn{1}{c}{\multirow{2}{*}{\makecell[c]{\\\textbf{Method}}}} &
  \multicolumn{3}{c}{\textbf{Classification (ROC AUC $\uparrow$)}} &
  \multicolumn{5}{c}{\textbf{Regression (RMSE $\downarrow$)}} &
  \multirow{2}{*}{\makecell[c]{\\\textbf{Avg. Rank}}} \\ \cmidrule(lr){2-4} \cmidrule(lr){5-9}
\multicolumn{1}{c}{} &
  \textbf{\makecell[c]{Homesite\\Insurance}} &
  \textbf{\makecell[c]{Ecom\\Offers}} &
  \textbf{\makecell[c]{HomeCredit\\Default}} &
  \textbf{\makecell[c]{Sberbank\\Housing}} &
  \textbf{\makecell[c]{Cooking\\Time}} &
  \textbf{\makecell[c]{Delivery\\ETA}} &
  \textbf{\makecell[c]{Maps\\Routing}} &
  \textbf{Weather} &  \\ \midrule

XGBoost\citep{xgboost}         & 0.9601 & 0.5763 & 0.8670 & 0.2419 & 0.4823 & 0.5468 & \textbf{0.1616} & 1.4671 & 4.4 \\
LightGBM\citep{lightgbm}        & 0.9603 & 0.5758 & 0.8664 & 0.2468 & 0.4826 & 0.5468 & 0.1618 & 1.4625 & 5.5 \\
CatBoost\citep{catboost}        & 0.9606 & 0.5596 & 0.8621 & 0.2482 & 0.4823 & \textbf{0.5465} & 0.1619 & 1.4688 & 5.6 \\ 
SNN\citep{SNN}            & 0.9492 & 0.5996 & 0.8551 & 0.2858 & 0.4838 & 0.5544 & 0.1651 & 1.5649 & 10.5 \\
DCNv2\citep{dcnv2}        & 0.9392 & 0.5955 & 0.8466 & 0.2770 & 0.4842 & 0.5532 & 0.1672 & 1.5782 & 11.4 \\
ResNet\citep{gorishniy2021revisiting}                & 0.9469 & 0.5998 & 0.8493 & 0.2743 & 0.4825 & 0.5527 & 0.1625 & 1.5021 & 8.0 \\
MLP-PLR\citep{mlp-plr}                & 0.9621 & 0.5957 & 0.8568 & 0.2438 & 0.4812 & 0.5527 & \textbf{0.1616} & 1.5177 & 4.4 \\
Trompt\citep{trompt}          & 0.9546 & 0.5792 & 0.8381 & 0.2596 & 0.4834 & 0.5563 & 0.1652 & 1.5722 & 11.0 \\
FT-Transformer\citep{gorishniy2021revisiting}  & 0.9622 & 0.5775 & 0.8571 & 0.2440 & 0.4820 & 0.5542 & 0.1625 & 1.5104 & 6.4 \\
TabR\citep{tabr}            & 0.9522 & 0.5850 & 0.8484 & 0.2851 & 0.4825 & 0.5541 & 0.1637 & \textbf{1.4622} & 8.6 \\
TabNet \citep{arik2021tabnet}         & 0.9531 & 0.5855 & 0.7701 & 0.2828 & 0.4813 & 0.5567 & 0.1651 & 1.5877 & 10.5 \\
TransTab\citep{wang2022transtab}        & 0.9564 & 0.5868 & 0.8498 & - & - & - & - & - & - \\
CM2\citep{cm2}             & 0.9560 & 0.5890 & 0.8392 & \textbf{0.2287} & 0.4838 & 0.5569 & 0.1638 & 1.5339 & 9.0 \\

TabPFN-2.5\citep{grinsztajn2026tabpfn25advancingstateart}             & 0.9443 & 0.5804 & - & 0.2356 & - & - & - & - &  -\\
\midrule
\textbf{\method}(Base) & \textbf{0.9635} & \textbf{0.6069} & \textbf{0.8698} & 0.2345 & \textbf{0.4806} & 0.5486 & 0.1618 & 1.4861 & \textbf{2.3} \\
\bottomrule
\end{tabular}
}
\end{table*}

\begin{table*}[h]
\centering
\small
\caption{
Performance comparison across models on the CreditRisk dataset.
}
\label{tab:performance_comparison}
\begin{tabular}{lccccc}
\toprule
\textbf{Model} & 
\textbf{Type} & 
\textbf{\#Params (Est.)} & 
\textbf{\#Features} & 
\textbf{ROC AUC $\uparrow$} &
\textbf{KS $\uparrow$} \\
\midrule
XGBoost\citep{xgboost}          & GBDT       & -   & 500  & 0.7080  & 0.3317 \\
LightGBM\citep{lightgbm}         & GBDT       & -   & 500  & 0.6949  & 0.3204 \\
TabNet\citep{arik2021tabnet}           & Seq-to-Seq & 29.71M   & 500   & 0.6581 & 0.2687 \\
MLP-PLR\citep{mlp-plr}           & Feedforward& 37.24M   & 500  & 0.6650  & 0.2860 \\
FT-Transformer\citep{gorishniy2021revisiting}   & Transformer & 31.52M   & 500   & 0.6832 & 0.2832 \\
CM2\citep{cm2}           & Transformer& 26.01M   & 500  & 0.6835  & 0.2922 \\
\midrule
MaskTab-Base     & Transformer& 25.16M   & 500   & 0.7149 & 0.3398 \\
MaskTab-L       & Transformer& 134.22M  & $2{,}000$  & \textbf{0.7584} & \textbf{0.4145} \\
MaskTab-Distill  & Transformer& 25.16M   & 500  & 0.7335  & 0.3802 \\
\bottomrule
\end{tabular}
\end{table*}

\subsection{Industrial‑scale data setting}\label{sec:data_setting}


We evaluate \method on tabular benchmarks that capture practical industrial constraints. Specifically, we use the public TabReD benchmark~\citep{rubachev2024tabredbenchmarktabularmachine}, which comprises eight datasets spanning three classification and five regression tasks, and a proprietary CreditRisk dataset containing \(6.4\times 10^{5}\) labeled instances and \(1.3\times 10^{7}\) unlabeled instances collected in 2024. For CreditRisk, we enforce a strictly temporal split to prevent information leakage: labeled data are partitioned by month into Training (Jan--Jun 2024; 330k), Validation (Jul--Sep 2024; 180k), and Test (Oct--Dec 2024; 130k). The dataset includes 2{,}500 features with pervasive missingness (49\% on average) and highly heterogeneous predictive strength (e.g., widely varying information value across features), closely mirroring production settings. Unlabeled samples used for self-supervised pre-training are drawn from the same time window as the labeled training split. Additional dataset details are provided in Appendix~\ref{appendix:dataset}.

\subsection{Experimental Settings}
\label{sec:exp-setting}





\noindent\textbf{Model Variants.} We define several \method variants for ablation, scaling, and deployment. \texttt{MaskTab-Base} serves as the default model, while \texttt{MaskTab-\{S/M/L/XL\}} scales capacity from 25M to 200M parameters. For production settings, we report \texttt{MaskTab-Distill}, a Base-sized student distilled from the best-performing \texttt{MaskTab-L}. See Appendix~\ref{appendix:model_variants} for details.

\noindent\textbf{Implementation Details.}
We train \texttt{MaskTab-Base} end-to-end with our hybrid supervised objective, following the optimization protocol of \citet{hoffmann2022training}. We use a batch size of 2048 and a cosine learning-rate schedule with 100 warmup steps, the learning rate peaks at \(1\times 10^{-4}\) and decays to \(1\times 10^{-5}\). To study scaling laws efficiently, we progressively scale along three axes and stop once performance saturates on each axis, substantially reducing training cost. When increasing the feature dimension, we add features in descending order of their Information Value (IV) scores. Models in our scaling study are trained with a two-stage schedule: (1) pre-training on large-scale labeled and unlabeled data with a learning rate scaled by model size, followed by (2) fine-tuning on labeled data only for stability. Finally, we conduct knowledge distillation by aligning the embedding representations of a high-performance \texttt{MaskTab-L} teacher model with those of a compact \texttt{MaskTab-Base} student model. All experiments were conducted using 8 NVIDIA A100 GPUs. Detailed hyperparameters, implementation procedures, and reproduction details for all baselines are provided in Appendix~\ref{appendix:implementation_details}.

\subsection{Baseline evaluation}

\noindent\textbf{Baseline.}
To thoroughly evaluate the effectiveness of our proposed method on the TabRed dataset, we benchmark it against two prominent categories of models: Gradient Boosted Decision Trees (GBDTs) and Deep Learning models, as shown in Table~\ref{tab:datasets}.

For the private dataset CreditRisk, we trained XGBoost\citep{xgboost} classifiers on each feature subset $\mathcal{G}_m$ ($d_m = 500m$, $m = 1, \dots, 5$). Models were fitted on $\mathcal{D}_{\text{train}}$, hyperparameters were selected via grid search on $\mathcal{D}_{\text{val}}$, and the best configurations were evaluated on the held-out $\mathcal{D}_{\text{test}}$, as detailed in Appendix~\ref{appendix:dataset}.

Figure~\ref{fig:ks_comp} shows diminishing returns once the feature count exceeds 1{,}500 (i.e., for $m \geq 3$), suggesting a dimensionality bottleneck. While XGBoost still gains slightly from additional features, the large train--test KS gap (Fig.~\ref{fig:ood_analysis}) indicates poor OOD generalization and potential performance decay after deployment. 

\noindent\textbf{Evaluation Protocol.}
We evaluate temporal robustness via out-of-time (OOT) testing by splitting the test set into monthly subsets. We report the mean Area Under the ROC Curve (AUC) and Kolmogorov-Smirnov (KS) across months to measure ranking quality and score stability under time shifts.

\subsection{Overall Performance}\label{sec:results}

\newcommand{\sameas}[1]{\textit{(same as \texttt{#1})}}

\begin{table}[t]
\centering
\small
\setlength{\tabcolsep}{5pt}
\caption{
Ablation study of MaskTab-Base components on the CreditRisk dataset.
}
\label{tab:ablation_study}
\begin{tabular}{lcccr}
\toprule
\textbf{Module Additions} & 
\textbf{ROC AUC $\uparrow$} & 
\textbf{KS $\uparrow$} \\
\midrule
Vanilla baseline  & 0.6472 &  0.2371    \\
+ Mask Embedding  & 0.6976 \textbf{{(+5.04\%)}}  &  0.3107 \textbf{{(+7.36\%)}}   \\
+ Twin Networks  & 0.7090 \textbf{{(+1.14\%)}}  &  0.3241 \textbf{{(+1.34\%)}}     \\
+ MoE  & 0.7149 \textbf{{(+0.59\%)}}  &  0.3398 \textbf{{(+1.57\%)}}     \\
\bottomrule
\end{tabular}
\end{table}

\begin{table}[t]
\centering
\small
\setlength{\tabcolsep}{5pt}
\caption{
Ablation study on shared vs.\ feature-specific mask embeddings on the CreditRisk dataset.
}
\label{tab:ablation_mask_embedding}
\begin{tabular}{lcc}
\toprule
\textbf{Configuration} & 
\textbf{ROC AUC $\uparrow$} & 
\textbf{KS $\uparrow$} \\
\midrule
Shared Mask Embedding   & 0.7149 & 0.3398 \\
Feature-Specific Mask Embedding & 0.7194 & 0.3387 \\
\bottomrule
\end{tabular}
\end{table}

\begin{table}[t]
\centering
\small
\setlength{\tabcolsep}{5pt}
\caption{Ablation study on imputation methods evaluated on the TabRed dataset.}
\label{tab:ablation_imputation}
\begin{tabular}{lcc}
\toprule
\textbf{Imputation Method} & 
\multicolumn{1}{c}{\textbf{HomeCredit}} & 
\multicolumn{1}{c}{\textbf{Sberbank}} \\
&
\multicolumn{1}{c}{\textbf{(AUC $\uparrow$)}} & 
\multicolumn{1}{c}{\textbf{(RMSE $\downarrow$)}} \\
\midrule
Zero Value       & 0.8625 & 0.2448 \\
Mode Value       & 0.8383 & 0.2419 \\
HyperImpute\citep{hyperimpute}      & {-}    & 0.2552 \\
ReMasker\cite{du2024remasker}     & 0.8609 & 0.2656 \\
Mask Embedding   & 0.8698 & 0.2345 \\
\bottomrule
\end{tabular}
\end{table}

The comparative results of \method and other methods on TabRed are presented in Table~\ref{tab:public_dataset_performance}. By calculating the average ranking across $8$ datasets, our method achieves an average rank of $\textbf{2.3}$. This performance surpasses XGBoost, the preeminent algorithm among GBDT methods, and significantly exceeds that of other deep tabular models. While some recent approaches, such as TabPFN-2.5~\citep{grinsztajn2026tabpfn25advancingstateart}, show strong performance on certain tabular benchmarks, they face scalability challenges on datasets exceeding 50k samples or 2k features. In contrast, MaskTab maintains competitive performance across diverse settings and scales efficiently to large real-world datasets.

Table~\ref{tab:performance_comparison} reports results on the CreditRisk dataset, demonstrating the effectiveness of the \method series. \texttt{MaskTab-Base}, trained solely on labeled data with a carefully selected set of interpretable features, significantly outperforms existing state-of-the-art methods. It achieves a 0.69\% improvement in OOT AUC and a 0.81\% improvement in KS over XGBoost. Our best-performing model from the extended experiments, \texttt{MaskTab-L}, shows an even greater improvement: a 5.04\% gain in AUC and an 8.28\% gain in KS over XGBoost, fully highlighting the advantage of large-scale representation learning. To balance performance and deployability, we apply knowledge distillation to transfer the representations learned by \texttt{MaskTab-L} into a compact student model, \texttt{MaskTab-Distill}. This variant shares the same architecture and input format as \texttt{MaskTab-Base}. It achieves a 2.55\% improvement in AUC and a 4.85\% improvement in KS over XGBoost, showing significant gains over the original base model while maintaining low inference latency and compatibility with standard interpretability tools such as SHAP. These results indicate that \texttt{MaskTab-Distill} is the recommended choice for deployment, achieving an optimal balance between accuracy, efficiency, and interpretability in real-world applications.

\subsection{Ablation Study}
To quantify the contribution of each design choice in \method, we perform ablations on industrial CreditRisk dataset by progressively adding components to a vanilla Transformer encoder baseline (zero imputation + task-only training).

As shown in Table~\ref{tab:ablation_study}, introducing learnable mask embeddings together with hybrid supervised pre-training (task loss + masked reconstruction) yields the largest improvement over the baseline (+5.04\% AUC, +7.36\% KS), highlighting the importance of explicitly modeling missingness and aligning pre-training with the downstream objective. Replacing joint training with the proposed twin-path training further improves performance (+1.14\% AUC, +1.34\% KS), confirming that separating reconstruction (masked input) from prediction (natural missingness) mitigates masking-induced shift. Finally, adding the MoE reconstruction head provides additional gains (+0.59\% AUC, +1.57\% KS), suggesting that expert routing improves reconstruction capacity via dynamic feature grouping. 

Table~\ref{tab:ablation_mask_embedding} compares shared versus feature-specific mask embeddings, where the negligible performance gap (AUC difference of 0.0045, KS difference of 0.0011), confirms that our feature-name embedding adequately captures inter-feature semantic distinctions, eliminating the need for dedicated per-feature initialization.

Table~\ref{tab:ablation_imputation} further validates the advantage of mask embeddings over conventional imputation strategies. Across both benchmarks, mask embeddings achieve the best AUC on HomeCredit (0.8698) and the lowest RMSE on Sberbank (0.2345), outperforming both heuristic methods (zero, mode) and learning-based alternatives (HyperImpute \citep{hyperimpute}, ReMasker \citep{du2024remasker}). This confirms that treating missingness as an explicit learnable signal yields more informative representations than substituting missing values with fixed or externally imputed ones.

\subsection{Scaling Law Analysis}\label{sec:scaling_analysis}




\begin{figure}[t]
    \centering
    \includegraphics[width=1.0\linewidth]{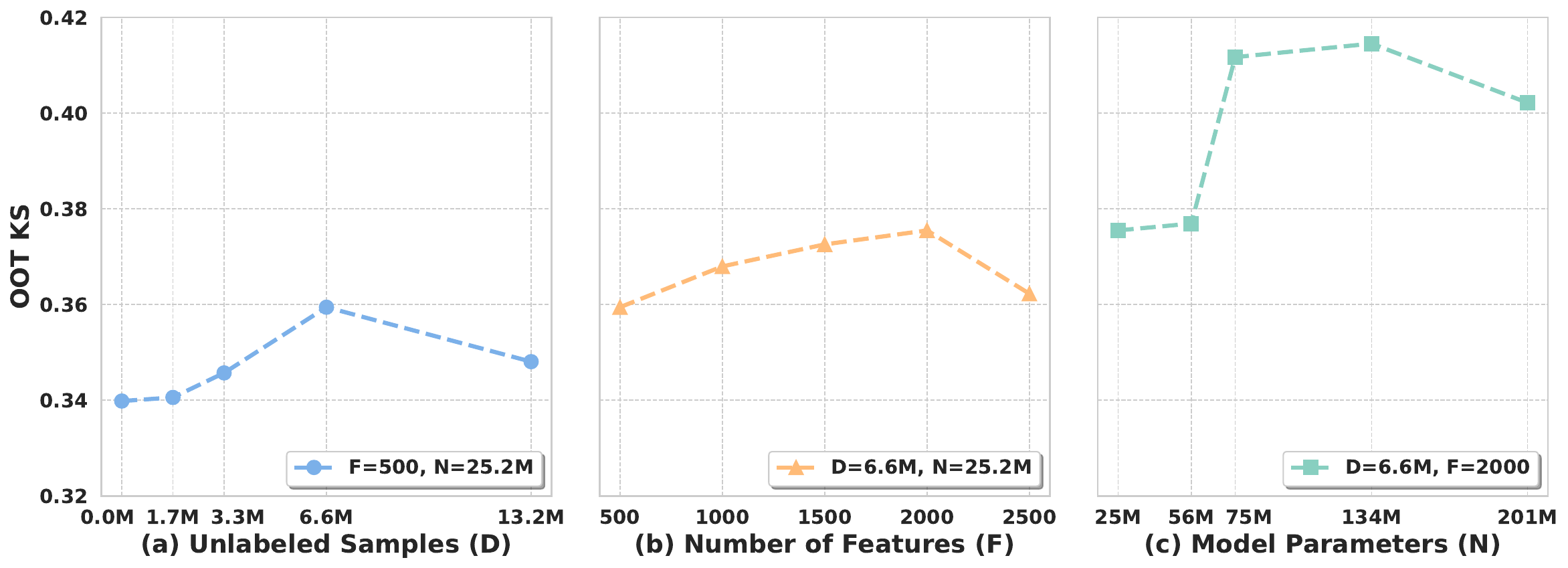}
    \caption{Scaling Law Analysis. (a) Scaling up the amount of unlabeled data, with the number of features fixed at 500 and the model parameters fixed at 25.2M. (b) Scaling up the number of features, with the amount of unlabeled data fixed at 6.6M and the model parameters fixed at 25.2M. (c) Scaling up the model parameters, with the amount of unlabeled data fixed at 6.6M and the number of features fixed at 2000.}
    \label{fig:scaling_law}
\end{figure}

\paragraph{Impact of the unlabeled data size.}
We first study data scaling by fixing the input feature budget to 500 and the model capacity to 25.2M parameters, and varying only the amount of unlabeled data. Concretely, we pre-train on the 330K labeled training split mixed with unlabeled samples at 0\(\times\), 5\(\times\), 10\(\times\), 20\(\times\), and 40\(\times\) the labeled set, and then fine-tune on labeled data only to ensure a fair downstream comparison. As shown in Figure~\ref{fig:scaling_law}, performance improves monotonically up to 20\(\times\) unlabeled data, where OOT KS reaches 0.3594 (+1.9\% over the labeled-only baseline). Scaling further to 40\(\times\) lowers KS to 0.3480, indicating diminishing returns and potential saturation under the current feature and capacity settings. In subsequent experiments, we therefore also examine scaling along the feature dimension and model size, which are complementary levers for better exploiting large unlabeled corpora.

\paragraph{Impact of the number of features.}

Following the data scaling study, we fix the unlabeled-to-labeled ratio at \(20\times\) and scale the feature dimension from 500 to 2{,}500 (step size 500), keeping all other settings unchanged. As shown in Figure~\ref{fig:scaling_law}, OOT KS increases with more features up to 2{,}000, peaking at 0.3754 (+1.6\% over 500 features), and then drops at 2{,}500 (0.3653), indicating saturation. Overall, \method effectively leverages additional high-dimensional, sparse features, demonstrating robustness on industrial tabular data with missingness and long-tailed feature distributions.

\paragraph{Impact of model parameters.}
The above results suggest diminishing returns when scaling either unlabeled data or feature dimension under a fixed architecture. We therefore scale model capacity under the best configuration (unlabeled-to-labeled ratio \(=20\times\), 2{,}000 features), increasing parameters from \texttt{MaskTab-Base} (25.16M) to \texttt{MaskTab-XL} (201.32M). Figure~\ref{fig:scaling_law} shows that KS improves monotonically up to \texttt{MaskTab-L} (134M), reaching 0.4145 (+3.91\% over the previous best 0.3754), but drops at \texttt{MaskTab-XL} (0.4021), again indicating saturation. Overall, \method exhibits consistent scaling behavior across data, features, and model size, providing a practical recipe for improving tabular models by jointly scaling these three factors.

\subsection{Knowledge Distillation for Deployment}
As shown in Figure~\ref{fig:ks_comp}, \texttt{MaskTab-L-2000} (134.21M) achieves the best KS among the \method variants, but is less suitable for production due to its high inference cost and limited interpretability with 2000 sparse features. We therefore distill \texttt{MaskTab-L-2000} into an efficient and auditable student, \texttt{MaskTab-Base-500}, by aligning their representations during training, yielding \texttt{MaskTab-Distill-500}. Distillation improves KS by 4.04\% over \texttt{MaskTab-Base-500}, and achieves a 9.3\(\times\) inference speedup while using only 500 interpretable features.

\begin{figure}[t]
    \centering
    \includegraphics[width=1.0\linewidth]{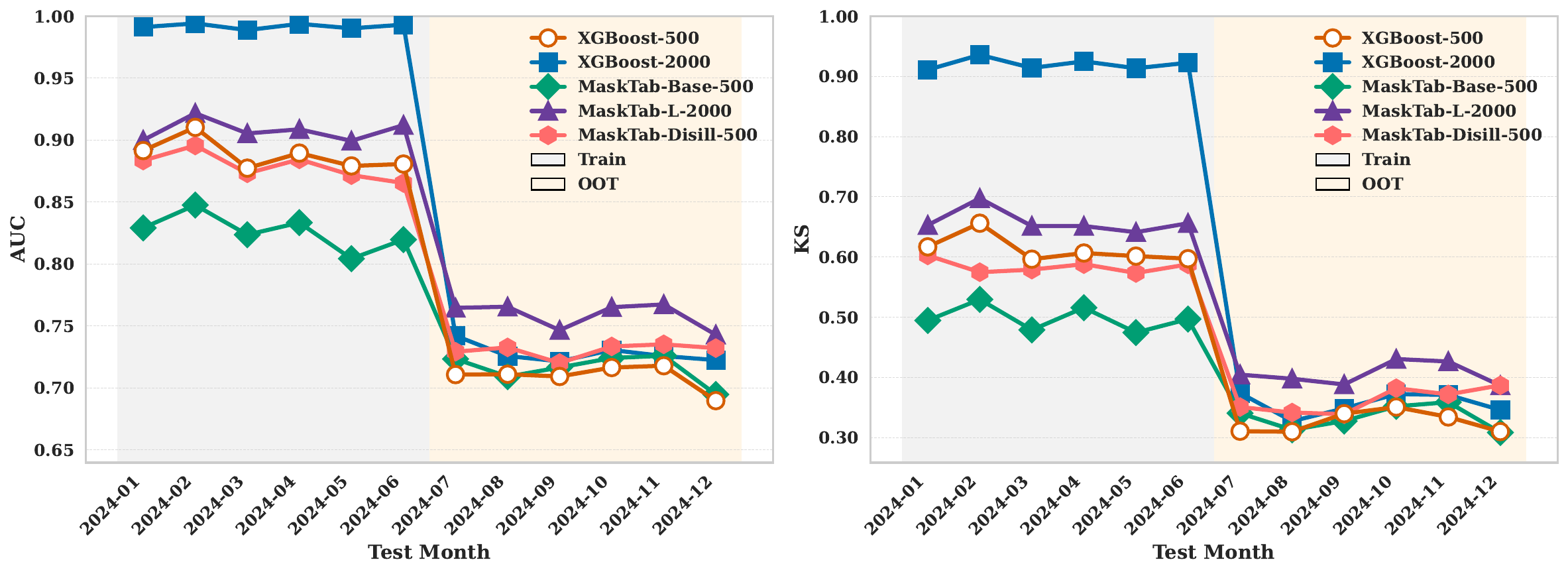}
    \caption{OOD Analysis. Monthly AUC (left) and KS (right) over time. Models are trained on 2024-01--2024-06 and evaluated on the OOT period 2024-07--2024-12; a smaller train--OOT gap indicates stronger robustness to distribution shift and better production generalization.}
    \label{fig:ood_analysis}
\end{figure}

\subsection{OOD Analysis}
Figure~\ref{fig:ood_analysis} compares training vs.\ out-of-time (OOT) KS to measure robustness under temporal distribution shift. GBDT baselines (e.g., XGBoost) exhibit a large generalization gap (training KS \(>\) 0.60 vs.\ OOT KS \(\approx\) 0.35), indicating substantial overfitting to period-specific signals. In contrast, \texttt{MaskTab-Base} markedly reduces this gap, and \texttt{MaskTab-Distill} further stabilizes it, yielding the smallest train--OOT discrepancy. These results suggest that our pre-training and distillation pipeline improves robustness to OOD drift, leading to more reliable performance after deployment.


\section{Conclusion}\label{sec:col}

In this paper, we present \method, a unified pre-training framework for industrial tabular data that are high-dimensional, riddled with missing entries, and rarely labeled at scale. \method uses learnable tokens to encode both masked and naturally missing values, and adopts twin-path hybrid pre-training that couples task supervision with masked reconstruction. An MoE reconstruction head further allocates capacity across heterogeneous features. Across large-scale industrial benchmarks, \method achieves state-of-the-art performance.

Importantly, we establish clear scaling trends: performance improves predictably as we scale unlabeled data, feature dimension, and model capacity, offering practical guidance for compute allocation. Finally, our distillation pipeline transfers these gains to compact, interpretable students for low-latency deployment.
\section*{Limitations}
Our scaling study is not exhaustive. Due to resource constraints, we adopt a greedy strategy that scales one dimension at a time (unlabeled data, feature count, or model size), which reveals consistent gains but does not yield a formal joint scaling law, a larger factorial exploration is needed. In addition, extending \method to richer modalities (e.g., more substantial text signals) and temporal/tabular time-series data remains future work.



\bibliography{custom}
\appendix
\section{Dataset Details}  \label{appendix:dataset}

\subsection{TabRed} 

\begin{table*}[h!]
\caption{
    The table presents details of nine datasets, eight from TabRed and one proprietary dataset named CreditRisk. For each dataset, we report the number of samples (\# samples), number of numerical features (\#num features), and number of categorical features (\#cat features). We also analyze the extent of missingness in each dataset by categorizing feature missing rates into three bins and reporting the proportion of samples falling into each bin.
}
\centering
\resizebox{\linewidth}{!}{
\begin{tabular}{@{}ccccccccc@{}}
\toprule
\multirow{2}{*}{\textbf{task type}} &
  \multirow{2}{*}{\textbf{dataset name}} &
  \multirow{2}{*}{\textbf{source}} &
  \multirow{2}{*}{\textbf{\#samples}} &
  \multirow{2}{*}{\textbf{\#num features}} &
  \multirow{2}{*}{\textbf{\#cat features}} &
  \multicolumn{3}{c}{\textbf{sample proportion at feature missing rates}} \\ \cmidrule(lr){7-9} 
                                &                    &         &     &    & & \textbf{(0\%, 33\%{]}} & \textbf{(33\%, 66\%{]}} & \textbf{(66\%, 100\%{]}} \\
                                \midrule
\multirow{3}{*}{Classification} & Ecom Offers (EO) & TabRed       & 160,057 & 113 & 6  & 0.00\%             & 0.00\%              & 0.00\%              \\
                                & Homesite Insurance (HI) & TabRed & 260,753 & 253 & 46 & 89.93\%       & 0.00\%              & 0.00\%               \\
                                & HomeCredit Default (HCD) & TabRed & 381,664 & 612 & 84 & 57.16\%       & 36.93\%        & 5.91\%          \\ 
                                & CreditRisk (CR) & Private & 13 million & 2475 & 25 & 6.75\%       & 89.17\%        & 4.08\%          \\
                                \midrule
\multirow{5}{*}{Regression}     & Sberbank Housing (SH) & TabRed  & 28,321  & 365 & 27 & 100.00\%        & 0.00\%              & 0.00\%               \\
                                & Cooking Time (CT) & TabRed       & 319,986 & 186 & 6  & 99.10\%       & 0.00\%              & 0.00\%               \\
                                & Delivery ETA (DE) & TabRed      & 350,516 & 221 & 2  & 91.50\%       & 3.71\%         & 0.00\%               \\
                                & Maps Routing (MR) & TabRed      & 279,945 & 984 & 2  & 97.08\%       & 2.82\%          & 0.10\%             \\
                                & Weather (W) & TabRed            & 423,795 & 100 & 3  & 0.00\%             & 2.21\%          & 0.03\%             \\ 
\bottomrule
\end{tabular}
}
\label{tab:datasets}
\end{table*}

We employed the TabReD benchmark, which exclusively consists of tabular data obtained from real-world industrial applications. This benchmark comprises $8$ datasets, including $3$ classification tasks and $5$ regression tasks. As outlined in Table~\ref{tab:datasets}, each dataset is characterized by a substantial number of samples and features. Furthermore, we evaluated the rates of missing features for each dataset and found that most exhibit varying degrees of incompleteness. Additionally, these datasets were divided into training, validation, and testing sets based on temporal criteria. These characteristics are frequently encountered in industrial applications, rendering this benchmark particularly suitable for assessing model performance in such contexts.

\subsection{CreditRisk}

\begin{figure}[h!]
\begin{center}
\includegraphics[width=0.85\linewidth]{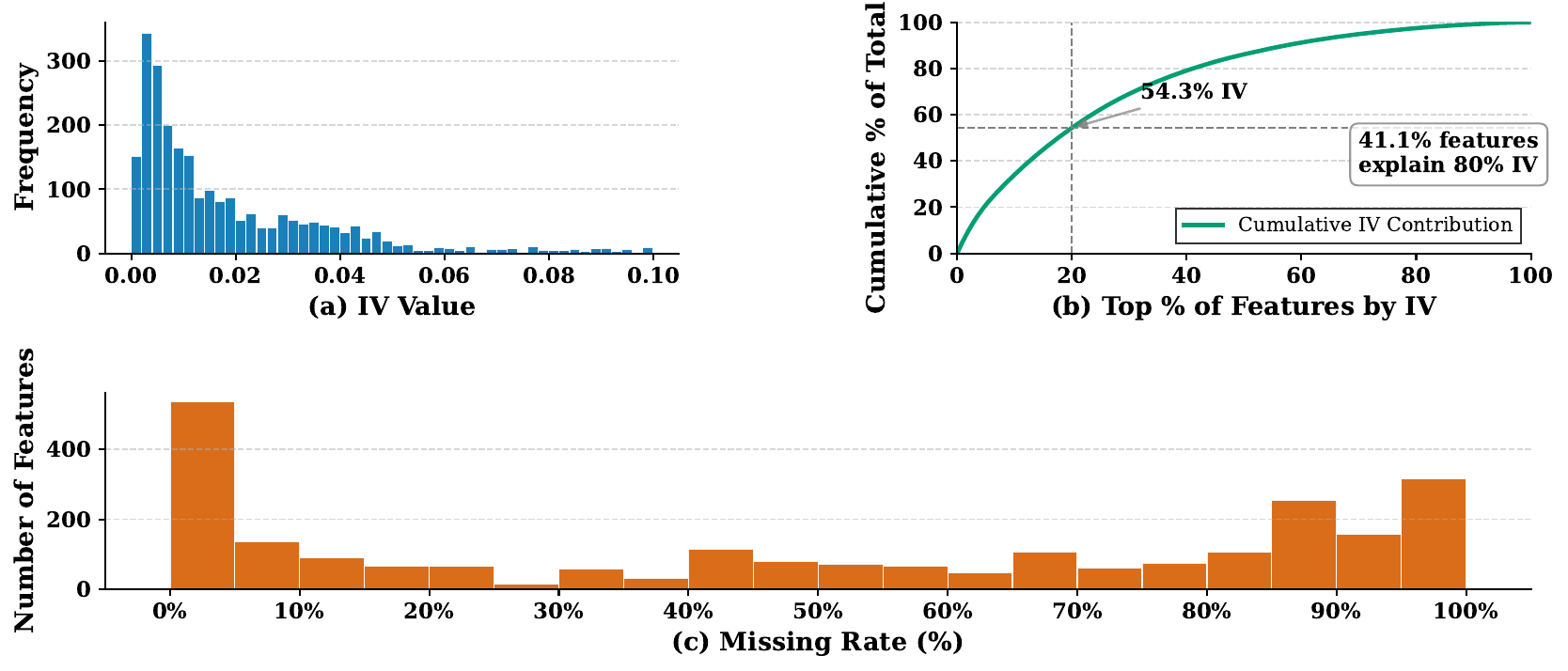}
\end{center}
\caption{Data Analysis of CreditRisk. Subfigure (a) displays the distribution of IV values across all features, showing a long-tail trend. Subfigure (b) illustrates the cumulative IV values as the feature proportion increases. Subfigure (c) presents the distribution of feature counts with varying missing rates.}
\label{fig:iv_missing}
\end{figure}

\noindent\textbf{Temporal Partitioning}. We utilize records spanning the entire year of 2024. Labeled data is partitioned monthly into: Training (Jan–Jun 2024): $\mathcal{D}_{\text{train}} = \left\{ (\mathbf{x}_i, y_i) \right\}_{i=1}^{N_{\text{train}}}$, 
$|\mathcal{D}_{\text{train}}| \approx 3.3 \times 10^{5}$. Validation (Jul–Sep 2024): $\mathcal{D}_{\text{val}} = \left\{ (\mathbf{x}_i, y_i) \right\}_{i=1}^{N_{\text{val}}}$, 
$|\mathcal{D}_{\text{val}}| \approx 1.8 \times 10^{5}$. Test (Oct–Dec 2024): $\mathcal{D}_{\text{test}} = \left\{ (\mathbf{x}_i, y_i) \right\}_{i=1}^{N_{\text{test}}}$, 
$|\mathcal{D}_{\text{test}}| \approx 1.3 \times 10^{5}$. Unlabeled data used for self‑supervised pre‑training are drawn from the same six‑month window as the training set, $\mathcal{D}_{\text{unsup}} = \left\{ \mathbf{x}_j \right\}_{j=1}^{M}$ with $M \approx 1.3 \times 10^{7}$, thereby eliminating any temporal leakage.

\noindent\textbf{Scale and dimensionality}. The total feature space contains $d=2{,}500$ attributes $(\mathcal F={f_k}_{k=1}^{d})$. With labeled instances numbering in the hundreds of thousands and unlabeled instances in the tens of millions, this dataset thus provides a platform for scaling up research.

\begin{table*}[t]
\centering
\small
\setlength{\tabcolsep}{6pt} 
\renewcommand{\arraystretch}{1.2} 
\caption{
Architecture and training configuration of the MaskTab model series.
All models share a Transformer backbone with progressively scaled capacity and input data.
}
\label{tab:hyperparams}
\begin{tabular}{lccccccr}
\toprule
\textbf{Model} & 
\textbf{\#Params(Non-Emb)} & 
\textbf{Layers} & 
\textbf{Number Heads} & 
\textbf{Key/Value Size} & 
$\textbf{d}_{\textbf{model}}$ \\ 
\midrule
MaskTab-Base     & $25.16M$ & 6 & 8 & 64 & 512   \\
MaskTab-S     & $56.62M$ & 6 & 12 & 64 & 768   \\
MaskTab-M     & $75.50M$ & 8 & 12 & 64 & 768   \\
MaskTab-L     & $134.22M$ & 8 & 16 & 64 & 1024   \\
MaskTab-XL     & $201.33M$ & 12 & 16 & 64 & 1024   \\
MaskTab-Distill     & $25.16M$ & 6 & 8 & 64 & 512   \\

\bottomrule
\end{tabular}
\end{table*}

\noindent\textbf{Pervasive missingness}. For each feature (k) we define a missing‑rate on the training set

\[ 
\begin{split}
\eta_k=\frac{1}{|\mathcal D_{\text{train}}|}\sum_{i=1}^{|\mathcal D_{\text{train}}|}\mathbb I!\bigl(v^{(i)}_{k}=\bot\bigr),\eta_k\in[0.0,1.0), 
\end{split}
\]
where $\bot$ denotes a missing value. The average missing rate across all features is $\bar\eta\approx0.49$, the distribution of ${\eta_k}$ is shown in Figure~\ref{fig:iv_missing} (distribution of feature missing rates).

\noindent\textbf{Feature‑selection protocol}. 
We adopt the conventional pipeline of ranking raw attributes by a univariate importance score $\mathcal{I}(f_k)$ (e.g., information value, IV). After sorting the $d = 2{,}500$ features in descending order of $\mathcal{I}$, we partition them into five incremental groups:

\[
\begin{split}
\mathcal{G}_m = \left\{ f_k \mid \operatorname{rank}(f_k) \leq m \times 500 \right\}, \\
m \in \{1,2,3,4,5\},
\end{split}
\]

so that the $m$-th group $\mathcal{G}_m$ contains the top $d_m = 500m$ features. This stratification serves two purposes. First, it provides a series of baseline models built on increasingly larger feature subsets, allowing us to quantify the marginal benefit of adding more variables. Second, by progressively introducing groups of lower‑importance features we can assess whether self‑supervised pre‑training amplifies the model’s ability to exploit weak signals.
The empirical distribution of IV values across all features is visualised in Figure~\ref{fig:iv_missing}.

\section{Model Variants Details} \label{appendix:model_variants}

\noindent\textbf{MaskTab-Base}. Trained on the \texttt{CreditRisk} dataset of interpretable features using labeled data and a mixed objective (e.g., reconstruction and classification). Serves as the baseline and student model.

\noindent\textbf{MaskTab-\{S/M/L/XL\}}. Scaled versions extending along model parameters, the number of features, and the volume of unlabeled data. \texttt{MaskTab-L} achieved the best performance among the scaled-up variants under the specific data and feature set used in this work. Pretrained on tens of millions of unlabeled samples, it was consequently employed as the teacher model to guide other variants via knowledge distillation. A summary of the hyperparameter settings for the \method model family is presented in Table~\ref{tab:hyperparams}.

\noindent\textbf{MaskTab-Distill}. A distilled version of \texttt{MaskTab-Base}. It retains the identical model architecture and parameter count as the \texttt{MaskTab-Base} but is trained to mimic the outputs and representations of the larger \texttt{MaskTab-L}. This process yields higher accuracy while preserving the base model's low latency and interpretability, making it ideal for production deployment.

\section{Implementation Details} \label{appendix:implementation_details}

\subsection{Model Comparison}
To situate our approach within the broader landscape of tabular data modeling, we compare \method with a range of representative baselines, including Gradient Boosted Decision Trees (GBDTs), deep learning models, and recent pre-training approaches, as detailed in Table~\ref{tab:public_dataset_performance} and Table~\ref{tab:performance_comparison}. 
All models are trained and evaluated on identical data splits. 


\renewcommand{\arraystretch}{1.0}
\begin{table}[h!]
\caption{
    The hyper-parameter optimization space for XGboost.
}
\centering
\resizebox{0.7\linewidth}{!}{
\scriptsize
\begin{tabular}{cc}
\toprule
\textbf{Hyper-parameter} & \textbf{Distribution} \\
\midrule
Learning rate & \{0.05, 0.1, 0.15, 0.2\} \\
N\_estimators & Const(1000) \\
Max depth & \{4, 5, 6, 7\} \\
Reg\_lambda & \{10, 50, 100, 500, 1000\} \\
Subsample & \{0.3, 0.5, 0.7, 0.9, 1.0\} \\
\bottomrule
\end{tabular}
}

\label{tab:Hyper_Parameter_xgboost}
\end{table}

\renewcommand{\arraystretch}{1.0}
\begin{table}[h!]
\caption{
    The hyper-parameter optimization space for LightGBM.
}
\centering
\resizebox{0.7\linewidth}{!}{
\scriptsize
\begin{tabular}{cc}
\toprule
\textbf{Hyper-parameter} & \textbf{Distribution} \\
\midrule
Boosting & \{gbdt, goss, rf, dart\} \\
Learning rate & \{0.05, 0.1, 0.15, 0.2\} \\
Num\_leaves & Const(31) \\
Min\_data\_in\_leaf & Const(20) \\
N\_estimators & Const(500) \\
Max depth & \{4, 5, 6, 7\} \\
Feature\_fraction & Const(1) \\
Max\_bin & Const(255) \\
Min\_data\_in\_bin & Const(3) \\
Bin\_construct\_sample\_cnt  & Const(200000) \\
\bottomrule
\end{tabular}
}

\label{tab:Hyper_Parameter_lightgbm}
\end{table}


The parameter search spaces for XGboost, and LightGBM are listed in Table~\ref{tab:Hyper_Parameter_xgboost}, and Table~\ref{tab:Hyper_Parameter_lightgbm}, respectively.

In addition to the TabReD benchmark, we have implemented some additional DNN-based methods, with the experimental settings for each method as follows:

\noindent\textbf{TabNet:}
 Use the official implementation\footnote{\url{https://github.com/dreamquark-ai/tabnet}}. The width of the decision prediction layer and the attention embedding for each mask is 8. The number of steps is 3, the learning rate is 0.02, and the early stopping patience is set to 20 out of a maximum of 200 epochs.

\noindent\textbf{TransTab:}
 Use the official implementation\footnote{\url{https://github.com/RyanWangZf/transtab}}. The model consists of 2 transformer layers. The dimensions of the hidden embedding and the feed-forward layers are 128 and 256, respectively, and the number of attention heads is 8. For training the downstream task, we use a batch size of 64, a learning rate of 1e-4, a dropout rate of 0, and an early stopping patience of 20 out of a maximum of 200 epochs.

\noindent\textbf{CM2:} 
 Use official implementation\footnote{\url{https://github.com/Chao-Ye/CM2}}. The token embedding dimension is 128, the hidden dimension of the feed-forward layers is 256, and the self-attention module has 8 heads. 3 transformer layers are used. It is trained with a batch size of 64, a learning rate of 1e-4, and a dropout rate of 0.15. The maximum training epoch is 200, and the patience value for early stopping is set to 10. We fine-tuned using pre-trained weights (\textit{CM2-v1}) and also fine-tuned from scratch, subsequently selecting the optimal results. 

\noindent\textbf{MaskTab-Base:} 
Trained end-to-end using a hybrid supervised strategy that combines pre-training and classification tasks. We adopt the optimization setup from \citet{hoffmann2022training}, using a batch size of 2048 and a cosine annealing schedule with a 100-step warmup. The learning rate peaks at $1\times10^{-4}$ and decays by a factor of 10. The model is evaluated after a single training stage without additional fine-tuning. All experiments were conducted using 8 NVIDIA A100 GPUs.

To quantify the contribution of each component in \texttt{MaskTab-Base}, we perform ablation studies by incrementally incorporating key modules: Parameter-Shared Mask Embedding, Siamese Network, and Scalable MoE Reconstruction Head.

\subsection{Scaling Laws.}
We empirically investigate scaling laws in tabular data learning by varying three factors: the amount of unlabeled data, the number of features, and the model parameters.

For data scaling, as introduced in Section~\ref{sec:data_setting}, we held constant a labeled training dataset of $3.3\times10^{5}$ samples while progressively increasing the volume of unlabeled data. This resulted in labeled-to-unlabeled data ratios of 1:5, 1:10, 1:20, and 1:40.

For feature scaling, we adhered to the experimental protocol outlined in Section~\ref{sec:data_setting}. Features are ranked by Information Value (IV) scores. We then expand the feature set in increments of 500 features up to 2500 features. This process mimics real-world scenarios where additional features typically have diminishing predictive power.

For model scaling, we introduce \texttt{MaskTab-\{S/M/L/XL\}}, a model family with increasing parameters. We adopt a two-stage training procedure to ensure stable learning across scales:
\begin{itemize}
    \item \textbf{Pretraining:} Mixed-supervision on both labeled and unlabeled data. The learning rate is scaled as:
    \[
    \text{LR}_{\text{scaled}} = \text{LR}_{\text{base}} \times \frac{1}{\sqrt{N/N_{\text{base}}}},
   \]
where $N$ is the parameter count of the current model and $N_{\text{base}}$ is that of \texttt{MaskTab-Base}.
    \item \textbf{Fine-tuning:} Supervised training on labeled data only, with learning rate reduced to $1\times10^{-5}$.
\end{itemize}

\subsection{Knowledge Distillation.}
To meet the real-time and interpretability constraints of risk control systems, we distill knowledge from a high-performance teacher model to a compact student. The teacher is a \texttt{MaskTab-L} model trained on 2,000 features, which was the best performer in our scaling-up experiments. The student is a \texttt{MaskTab-Base} model restricted to 500 interpretable features. The distillation process first caches the embedding vectors generated by the teacher model on the training set. It then trains the student model using these cached embeddings as soft targets, with a distillation loss enforcing representation alignment (Section~\ref{sec:distillation}).

\end{document}